**Validation of a Small Language Model for**

**DSM-5 Substance Category Classification in Child Welfare Records**

Brian E. Perron [a] , Dragan Stoll [b,c], Bryan G. Victor [d], Zia Qi[a], Andreas Jud [b,d], Joseph P. Ryan[a]

[a]School of Social Work, University of Michigan

[b, c] Institute of Psychology and Education, Ulm University, Germany

[c] School of Social Work, ZHAW Zurich University of Applied Sciences, Switzerland

[d] Child and Adolescent Psychiatry, Psychosomatics, and Psychotherapy, Ulm University Clinics, Ulm, Germany

Brian E. Perron, beperron@umich.edu, https://orcid.org/0009-0008-4865-451X

Dragan Stoll, dragan.stoll@zhaw.ch, https://orcid.org/0009-0008-9088-5282

Bryan Victor: bvictor@wayne.edu, https://orcid.org/0000-0002-2092-912X

Zia Qi, qizixuan@umich.edu, https://orcid.org/0000-0002-8407-0465

Andreas Jud, andreas.jud2@zhaw.ch, https://orcid.org/0000-0003-0135-4196

Joseph P. Ryan, joryan@umich.edu

**Abstract**

**Background:** Recent studies have demonstrated that large language models (LLMs) can perform binary classification tasks on child welfare narratives — detecting the presence or absence of constructs such as substance-related problems, domestic violence, and firearms involvement. However, whether smaller locally deployable models can move beyond binary detection to classify specific substance types from these narratives remains untested.

**Objective:** Validate a locally hosted LLM classifier for identifying specific substance types aligned with DSM-5 categories in child welfare investigation narratives.

**Methods:** A locally hosted 20-billion-parameter LLM classified child maltreatment investigation narratives from a Midwestern U.S. state. Records previously identified as containing substance-related problems were passed to a second classification stage targeting seven DSM-5 substance categories. Expert human review of 900 stratified cases assessed classification precision, recall, and inter-method reliability (Cohen's kappa). Test-retest stability was evaluated using approximately 15,000 independently classified records.

**Results:** Five substance categories achieved almost perfect inter-method agreement ($\kappa$ = 0.94–1.00): alcohol, cannabis, opioid, stimulant, and sedative/hypnotic/anxiolytic. Classification precision ranged from 92% to 100% for these categories. Two low-prevalence categories (hallucinogen, inhalant) performed poorly. Test-retest agreement ranged from 92.1% to 99.1% across the seven categories.

**Conclusions:** A small, locally hosted LLM can reliably classify substance types from child welfare administrative text, extending prior work on binary classification to multi-label substance identification.

**Validation of a Small Language Model for**

**DSM-5 Substance Category Classification in Child Welfare Records**

Child welfare agencies generate vast amounts of administrative data through routine operations. Among the most information-rich elements of these records are free-text narratives (e.g., investigation summaries, dispositional findings, and case notes) in which caseworkers document the circumstances surrounding allegations of child maltreatment. These narratives frequently contain detailed references to parental and caregiver substance use, including specific substances, patterns of use, drug test results, and substance-related behaviors observed during investigations (Canfield et al., 2017; Henry et al., 2020; Kepple, 2018). Despite the richness of this information, these narratives remain largely inaccessible for systematic analysis. Administrative databases typically capture substance involvement, if at all, as a single binary indicator, collapsing the complexity of substance-specific patterns into an undifferentiated flag (Victor et al., 2018).

This limitation has practical consequences. The landscape of substance use affecting families in the child welfare system has shifted dramatically over the past two decades. The opioid epidemic in the United States drove sharp increases in child welfare caseloads and foster care entries during the 2010s (Radel et al., 2018; Quast et al., 2018). Findings from Ohio indicate that while opioid use among child welfare-involved families has declined in recent years, cannabis use has remained relatively common and stable (Ye et al., 2025). Yet, cannabis legalization across states has introduced new questions about how changing legal status will and should affect child welfare decision-making (Cerda et al., 2020; Ryan et al., 2025). Alcohol use disorder remains the most prevalent substance-related problem among parents in the child welfare population (Lipari & Van Horn, 2017; Ogbonnaya et al., 2019), though it has figured less prominently in recent policy responses.

In administrative systems that do not distinguish among substances in their structured records, agencies and researchers are unable to track these divergent trends, assess substance-specific risks, or evaluate the effectiveness of targeted interventions. Promisingly however, recent work has established that large language models (LLMs) can reliably perform binary classification on child welfare investigation narratives. Perron et al. (2024) demonstrated that locally hosted LLMs achieved classification performance comparable to human experts when detecting the presence or absence of substance-related problems in investigation summaries, with F1 scores of 89–95% across four open-source models.

Qi et al. (2026) extended this work through a systematic benchmarking framework, showing that small reasoning-enabled models achieved almost perfect agreement on binary classification tasks (substance-related problems, firearms, and opioids) and substantial agreement on the more complex task of domestic violence detection. Stoll et al. (2025) further demonstrated that LLMs can classify child maltreatment subtypes, with model performance exceeding inter-rater agreement among human reviewers on multi-class classification. Together, these studies establish that LLMs, including small, locally deployed models, can reliably detect the presence or absence of key constructs in child welfare narratives.

What remains untested is whether the foundational language capability of a small local model extends beyond binary detection to the more demanding task of classifying specific substance types. The challenge is significant, given that child welfare narratives exhibit a considerable range of language patterns for substance-related content that must be interpreted in context. A reference to alcohol, for example, might appear as "father was intoxicated at the time of the home visit," "mother reports father drinks a six-pack every night," "CPS worker noted the smell of alcohol on the caregiver's breath," "parent denies any alcohol use," or "family was referred for alcohol assessment but declined services." Each phrasing requires contextual interpretation to determine whether an active alcohol problem is present.

This complexity, which multiplies across substance types, is a particular challenge for smaller locally deployed models. Frontier commercial models such as ChatGPT (OpenAI), Claude (Anthropic), and Gemini (Google) operate with hundreds of billions of parameters, sometimes approaching or exceeding one trillion parameters, and benefit from massive training corpora and extensive fine-tuning. Smaller open-source models with 4 to 70 billion parameters, which are practical for local deployment on agency-controlled hardware, necessarily encode less linguistic knowledge and may struggle with the breadth of vocabulary and the contextual reasoning required for fine-grained substance classification. The question is whether the smaller models retain sufficient foundational language capability to make reliable substance-type distinctions from the language patterns found in child welfare narratives.

The binary task of identifying substance-related problems has been validated in previous work (Perron et al., 2024; Qi et al., 2026). This study extends those results to a more nuanced classification. Specifically, among records identified as having a problem related to substances, we evaluate whether a small local model can reliably identify which specific substances are involved, classifying across seven categories aligned with DSM-5 substance use disorder groupings (American Psychiatric Association, 2013).

## Methods

### Data Source

Investigation narratives were drawn from child welfare dispositional findings in a single Midwestern U.S. state, sampled from multiple years (2013-2024) of child protective services investigations. Dispositional findings are standardized narrative summaries documenting the circumstances, evidence, and outcomes of each investigation. Each record includes an unstructured narrative text field and structured fields for the investigation outcome. Records were sampled across years for validation purposes. This study was approved by the authors' institutional review board (IRB).

### Classification Pipeline

Classification used a two-stage approach. In Stage 1, investigation narratives were evaluated for the presence of substance-related problems (SRP) using a binary classification prompt. As noted, this stage has been validated in prior research in prior work and is not the focus of the present study. In Stage 2, the focus of this study, narratives classified as having an SRP present underwent substance-specific classification across seven categories aligned with DSM-5 substance use disorder groupings: (1) alcohol, (2) cannabis, (3) opioid, (4) stimulant, (5) sedative/hypnotic/anxiolytic, (6) hallucinogen, and (7) inhalant.

A single prompt classified each narrative across all seven DSM-5 substance categories simultaneously, including category definitions, classification rules, and the required output schema (see Appendix A). Categories were not mutually exclusive. For each substance identified in the narrative, the model returned the classification and verbatim text extracts supporting the decision.

The classification model was gpt-oss:20b, a 20-billion-parameter open-source LLM quantized to 4-bit precision (OpenAI, 2025), and was deployed locally on dual NVIDIA A6000 graphics processing units (48 GB VRAM each). All processing occurred on-premises with no data transmitted to external servers, preserving the confidentiality of child welfare records. Inference was performed with a temperature parameter of 0.2 to promote classification consistency and an 8,192-token context window to accommodate lengthy narratives.

**Validation Design**

Validation employed stratified random sampling across three strata. First, for each of the seven substance categories, 100 records classified by the model as positive were randomly selected for human review, yielding 700 substance-specific validation cases for precision assessment. Second, 100 records classified as SRP-absent were sampled to assess recall through false-negative analysis. Third, 100 SRP-present records that received no substance-specific classification were sampled to evaluate the appropriateness of unclassified cases. The total validation sample comprised 900 records.

A human reviewer assessed each case using the full dispositional narrative and compared the result with the model's classification. For precision assessment, the reviewer determined whether each positive classification was correct given the narrative content. For the extraction assessment, each text phrase extracted by the model was evaluated for relevance to the target substance classification. An automated step assessed whether extracted phrases appeared verbatim in the source text using case-insensitive string matching.

For recall assessment, the 100 SRP-absent cases were prescreened using keyword-based regular expressions covering comprehensive substance terminology (drug names, street terms, behavioral indicators) to identify potential false negatives before human review.

**Inter-Method Reliability**

Agreement between LLM classification and expert human review was quantified using Cohen's kappa (κ; Cohen, 1960), with interpretation following the Landis and Koch (1977) benchmarks: slight (0.00–0.20), fair (0.21–0.40), moderate (0.41–0.60), substantial (0.61–0.80), and almost perfect (0.81–1.00). Because kappa can be affected by prevalence imbalance in the validation sample, we also report the prevalence-adjusted bias-adjusted kappa (PABAK; Byrt et al., 1993). Ninety-five percent confidence intervals for precision estimates were calculated using the Wilson score method (Wilson, 1927).

**Test-Retest Stability**

Because LLMs are probabilistic, generating outputs through token-by-token sampling rather than deterministic rule application, classification consistency across independent runs is not guaranteed, even with identical inputs and low temperature settings. To assess classification stability, approximately 15,000 investigation records that appeared in two independent data extractions were classified in separate processing runs using identical prompts and model parameters. These overlapping records provided a natural test-retest assessment of classification reproducibility. Agreement was calculated as the percentage of records receiving concordant classifications across both runs.

**Results**

**Inter-Method Reliability and Classification Precision**

Table 1 presents the consolidated validation results for all seven substance categories. Five categories demonstrated almost perfect agreement between LLM classification and expert human review. Alcohol and opioid achieved perfect precision (100%) with κ = 1.00. Cannabis and stimulants achieved near-perfect precision (99.0%) with κ = 0.99. Sedative/hypnotic/anxiolytic demonstrated strong precision (92.0%) with κ = 0.94, also in the almost perfect range. PABAK values closely tracked Cohen's κ across all categories, confirming that prevalence imbalance in the validation sample did not substantially distort agreement estimates. Two categories performed poorly. Hallucinogen precision was 56.1% (κ = 0.63, substantial agreement), and inhalant precision was 35.0% (κ = 0.42, moderate agreement). These categories are reported below the line in Table 1 and excluded from substantive application.

[INSERT TABLE 1 ABOUT HERE]

**Extraction Quality**

For each of the 700 substance-specific validation cases, the model extracted verbatim text phrases from the source narrative as evidence for its classification. Across these cases, the model produced 1,412 text extractions in total (averaging approximately two per positive classification). Human review determined that 90.5% (1,278/1,412) of these extractions were valid and accurately represented substance-related content for the target category. Extraction precision exceeded 94% for the four highest-performing categories (alcohol, cannabis, opioid, stimulant). Automated string matching confirmed that 92.8% of all extracted phrases appeared verbatim in the source text. The remaining 7.2% were semantically accurate paraphrases in which the model slightly reworded source content while preserving meaning.

A third validation stratum examined 100 SRP-present records that the model classified as substance-involved but assigned no specific substance category. Human review determined that 92% of these contained only generic substance references (e.g., "substance abuse," "drug use," "addiction") without naming a specific substance — the model correctly identified substance involvement but lacked a specific substance to classify. The remaining 8% represented edge cases in which the narrative language was sufficiently ambiguous that an expert and a model could reasonably disagree on whether a substance-related problem was present.

**Test-Retest Stability**

Table 2 presents test-retest agreement for approximately 15,000 records classified independently in two separate processing runs. SRP classification demonstrated 99.1% agreement. Among substance-specific categories, agreement ranged from 92.1% (stimulant) to 97.1% (alcohol, cannabis). The small variation across runs reflects the probabilistic nature of LLM inference, in which token sampling can produce different outputs for identical inputs. The high agreement rates indicate that the low-temperature setting (0.2) effectively constrained variability in the substance classification task.

[INSERT TABLE ABOUT HERE]

## Discussion

This study demonstrates that a 20-billion-parameter locally hosted LLM can reliably classify specific substance types from free-text child welfare investigation narratives. Five of seven DSM-5-aligned substance categories achieved almost perfect inter-method agreement between LLM classification and expert human review. Test-retest agreement across independent processing runs were extremely high. These results extend prior findings, demonstrating that a small local model retains sufficient foundational language capability for this purpose.

The fact that a 20-billion-parameter model, roughly two orders of magnitude smaller than frontier commercial models, achieved near-perfect precision for five substance categories suggests that the linguistic knowledge required for this task is well within the capability of models practical for local agency deployment. This is a meaningful finding for the field, as it confirms that agencies need not rely on cloud-based commercial services, with their attendant confidentiality risks and per-token costs, to obtain substance-specific classifications from their administrative records.

**Poor Performance for Hallucinogen and Inhalant Categories**

Two substance categories fell well below acceptable thresholds: hallucinogen (precision 56.1%) and inhalant (precision 35.0%). Two interacting factors likely account for this pattern. First, the terminology associated with these substance categories overlaps substantially with non-substance language common in child welfare narratives. Inhalant-related terms such as "gas," "spray," "paint," and "cleaning products" appear frequently in descriptions of home environments, household hazards, and child safety concerns unrelated to substance use. The model must disambiguate these references from genuine inhalant abuse based on context alone, a task that proves error-prone.

Examination of misclassified cases reveals distinct error patterns. In one case, the model classified a narrative as inhalant-positive based on the phrase "tubes were for inhaling or snorting drugs," when the tubes in question were paraphernalia for snorting cocaine and heroin; the model confused the route of administration (nasal inhalation) with the substance category (inhalants), when the correct classifications were stimulant and opioid. Similarly, a hallucinogen false positive was triggered by the phrase "acid used to break down crack cocaine for injection," in which "acid" referred to a chemical solvent used in stimulant preparation rather than to LSD.

In both instances, the model matched on a category-associated term without resolving the surrounding context. Notably, when contextual signals are unambiguous, the model classifies correctly; for example, a narrative describing a father found unconscious with keyboard cleaner spray, one of the most commonly abused inhalant products, (Perron & Dimit, under review), was accurately identified as inhalant use. The difficulty is not that the model lacks the capacity to recognize these substances, but that the terminology requires contextual disambiguation that the model performs unreliably.

By contrast, the high-performing categories rely on terminology that is largely substance-specific (e.g., "methamphetamine," "heroin," "Suboxone"), requiring far less contextual disambiguation. Second, both hallucinogen and inhalant use are among the least prevalent substance categories in the child welfare population. When base rates are very low, even a modest number of contextual misclassifications substantially degrades precision, because false positives constitute a disproportionate share of all positive classifications. These factors compound: the categories with the most ambiguous terminology are also the rarest, creating conditions under which disambiguation errors have the greatest impact on measured precision. Improving performance in these categories would likely require enhanced prompt engineering, extensive exemplars, post-classification filtering rules targeting known ambiguous terms, or all of the above. In the interim, these categories should be excluded from analyses requiring high precision, as we have done in presenting our results.

**Edge Cases and Classification Ambiguity**

The small residual error in both SRP-level (Stage 1) and substance-type classifications (Stage 2) reflects genuine ambiguity in the source narratives rather than systematic model failure. Child welfare investigation summaries are written by caseworkers under time pressure, and the language used to document substance involvement varies considerably in clarity and specificity. Some narratives contain clear, unambiguous references ("father tested positive for methamphetamine"); others use hedged or second-hand language ("there were concerns expressed by the neighbor about possible drug use in the home"). In the most ambiguous cases, an expert and a model can reasonably disagree on whether a substance-related problem is present and, if so, which substance is involved.

This ambiguity establishes a natural ceiling on classification performance that no method, whether human or automated, can exceed. Inter-rater reliability studies across clinical and social service contexts consistently demonstrate that trained professionals disagree with one another at non-trivial rates on classification tasks requiring interpretive judgment (e.g., Stoll et al., 2025). If two qualified human reviewers would not achieve perfect agreement on whether a given narrative indicates an active alcohol problem versus generic "substance abuse," then holding an automated classifier to a standard of perfect agreement with a single reviewer imposes an unrealistic benchmark. The relevant question is not whether the model achieves perfect accuracy but whether it performs within the range of disagreement that would be observed between qualified human raters.

**Opportunities for Administrative Data Enrichment**

The validated classification pipeline transforms existing administrative records into a source of substance-specific data without requiring changes to data collection practices, imposing any additional burden on caseworkers, or transmitting confidential information outside secure environments. Population-level surveillance becomes possible: agencies can monitor substance-specific trends over time using data already in their records. Longitudinal analysis of how the substance landscape has shifted, from the opioid crisis of the mid-2010s to subsequent stimulant surges, can be conducted retrospectively. Substance-specific granularity enables more targeted questions than binary indicators allow, such as whether service provision aligns with the specific substances affecting families in a given region.

Beyond surveillance, substance-specific classifications can be linked to existing structured fields in administrative databases to support analytic questions that were previously inaccessible. For example, agencies could examine whether investigation outcomes, service referrals, or placement decisions differ as a function of the specific substance involved, controlling for other case characteristics. Researchers could investigate whether families affected by opioid use follow different child welfare trajectories than those affected by alcohol or stimulants. Because the classification pipeline processes historical records, these analyses can be applied to years of accumulated data without prospective data collection, effectively enriching administrative datasets that were never designed to capture substance-specific information. The result is a low-cost, scalable method for converting existing unstructured text into structured analytic variables that extend the research and practice utility of records agencies already maintain.

**Limitations**

Several limitations should be noted. First, recall was assessed at the SRP level only; per-substance recall (the rate at which the classifier misses specific substance types among SRP-present records) was not independently evaluated. Second, validation samples of 100 cases per category provide reasonable but not definitive estimates; rare edge cases may be underrepresented. Third, these results are based on a single state's records and may not generalize to jurisdictions with different documentation practices, terminology, or populations.

**Conclusion**

This study validates a locally hosted LLM-based classification pipeline for identifying specific substance types in child welfare investigation narratives. Five DSM-5-aligned substance categories — alcohol, cannabis, opioid, stimulant, and sedative/hypnotic/anxiolytic — achieved almost perfect inter-method reliability with expert human review, classification precision, and test-retest agreement. These findings extend prior work on binary substance detection to multi-class substance-type identification, demonstrating that a small, locally deployed model has the foundational language capability required for this more demanding classification task.

**Declaration of Generative AI and AI-Assisted Technologies in the Writing Process**

During the preparation of this work, the authors used Claude (Anthropic) in order to assist with manuscript editing and with the construction of code used in the classification pipeline. After using this tool, the authors reviewed and edited the content as needed and take full responsibility for the content of the publication.

**Table 1**

*Validation Results by Substance Category*

| Substance | *n* | Precision % (95% CI) | κ (95% CI) | PABAK | Interpretation | Extraction Precision % |
|---|---|---|---|---|---|---|
| Alcohol | 100 | 100.0 (96.3–100.0) | 1.00 (1.00–1.00) | 1.00 | Almost Perfect | 99.3 |
| Cannabis | 100 | 99.0 (94.6–99.8) | 0.99 (0.98–1.00) | 0.99 | Almost Perfect | 98.9 |
| Opioid | 100 | 100.0 (96.3–100.0) | 1.00 (1.00–1.00) | 1.00 | Almost Perfect | 98.1 |
| Stimulant | 100 | 99.0 (94.6–99.8) | 0.99 (0.98–1.00) | 0.99 | Almost Perfect | 94.6 |
| Sedative/Hypnotic /Anxiolytic | 100 | 92.0 (85.0–95.9) | 0.94 (0.90–0.98) | 0.95 | Almost Perfect | 92.8 |
| Hallucinogen | 98 | 56.1 (46.3–65.5) | 0.63 (0.53–0.73) | 0.71 | Substantial | 75.5 |
| Inhalant | 100 | 35.0 (26.4–44.7) | 0.42 (0.29–0.54) | 0.57 | Moderate | 58.6 |

*Note.* Each substance category was validated by comparing expert assessments with model classifications on 100 randomly sampled positive cases and 200 negative cases (100 SRP-absent, 100 SRP-present with no substance classification). κ = Cohen's kappa. PABAK = prevalence-adjusted bias-adjusted kappa. Extraction precision based on 1,412 text phrases extracted across the 700 substance-specific validation cases (see Extraction Quality section). Verbatim extraction match rate = 92.8%.

**Table 2**

*Test-Retest Reliability*

| Classification | Both Present | Both Absent | Disagree | % Agreement |
|---|---|---|---|---|
| SRP | 7,707 | 7,496 | 136 | 99.1 |
| Alcohol | 2,272 | 5,346 | 89 | 97.1 |
| Cannabis | 4,258 | 3,358 | 91 | 97.1 |
| Opioid | 1,385 | 6,072 | 250 | 95.1 |
| Stimulant | 1,647 | 5,576 | 484 | 92.1 |

*Note.* SRP agreement is calculated across all overlapping records. Substance-specific

agreement is calculated among records classified as SRP-present in both processing runs.

Disagree = sum of discordant classifications (present in one run, absent in the other). Both runs used identical prompts, model parameters, and infrastructure.

**Appendix A: Classification Prompt**

# Substance Class Detection Prompt - Multilabel Classification with Evidence Extraction

## Task Description
You are analyzing a child welfare investigation summary that has been identified as involving substance abuse. Your task is to:
1. Classify which specific substance classes are mentioned or implicated in this case
2. Extract the exact text mentions that support each classification

Task: For each of the 7 substance classes below, classify whether that substance class is "Present" or "Absent", AND provide the exact phrases/words from the text that led to your classification.

---

## SUBSTANCE CLASS DEFINITIONS

### 1. ALCOHOL
Includes any mention of:
- **Beverages**: Beer, wine, liquor, spirits, hard alcohol, mixed drinks, moonshine
- **Behaviors**: Drinking, intoxicated, drunk, inebriated, under the influence of alcohol, alcohol abuse, alcoholism, alcohol dependence, binge drinking, blackouts
- **Evidence**: Smell of alcohol, alcohol on breath, open containers, DUI/DWI, alcohol treatment, AA/Alcoholics Anonymous, detox for alcohol
- **Slang**: Booze

### 2. CANNABIS
Includes any mention of:
- **Forms**: Marijuana, hashish, hash oil, THC, CBD with intoxicating effects, edibles, wax, dabs, concentrates, vape cartridges with THC
- **Behaviors**: Smoking marijuana, getting high on weed, cannabis use disorder
- **Evidence**: Positive THC test, marijuana paraphernalia (pipes, bongs, rolling papers), smell of marijuana, marijuana plants
- **Slang**: Weed, pot, grass, herb, bud, ganja, dope, reefer, 420, blunt, joint, dab, kush

### 3. HALLUCINOGEN

Includes any mention of:
- **Classic hallucinogens**: LSD (acid), psilocybin (magic mushrooms, shrooms), mescaline, peyote, DMT, ayahuasca
- **Dissociatives**: PCP (angel dust, phencyclidine), ketamine (Special K), DXM (dextromethorphan, robotripping, triple C)
- **Other**: MDMA/ecstasy/molly (when used for hallucinogenic effects), salvia, NBOMe (N-bomb)
- **Behaviors**: Tripping, hallucinating from drug use, psychedelic use
- **Evidence**: Positive PCP test, hallucinogen paraphernalia

### 4. INHALANT

Includes any mention of:
- **Volatile solvents**: Paint thinner, gasoline, glue, felt-tip marker fluid, correction fluid, lighter fluid
- **Aerosols**: Spray paint, deodorant sprays, hairspray, cooking sprays, fabric protector sprays
- **Gases**: Nitrous oxide (whippets, laughing gas), butane, propane, refrigerant gases, ether, chloroform
- **Nitrites**: Amyl nitrite, butyl nitrite (poppers, rush, snappers)
- **Behaviors**: Huffing, sniffing, bagging, inhaling fumes to get high, dusting
- **Evidence**: Chemical burns around mouth/nose, rags or bags with chemical smell

### 5. OPIOID

Includes any mention of:
- **Prescription opioids**: Oxycodone (OxyContin, Percocet, Roxicodone), hydrocodone (Vicodin, Norco, Lortab), morphine, codeine, fentanyl (Duragesic, Actiq), methadone, buprenorphine (Suboxone, Subutex), tramadol (Ultram), hydromorphone (Dilaudid), meperidine (Demerol)
- **Illicit opioids**: Heroin (smack, H, dope, black tar, china white), illicit fentanyl, carfentanil
- **Treatment medications**: Methadone clinic, Suboxone treatment, MAT (medication-assisted treatment) for opioids, naloxone/Narcan use or possession
- **Behaviors**: Opioid use disorder, opioid addiction, nodding off, overdose from opioids
- **Evidence**: Positive opioid test, track marks, needle marks, opioid paraphernalia (syringes, spoons, tourniquets)

- **Slang**: Oxy, percs, lean, purple drank, vikes

### 6. SEDATIVE/HYPNOTIC/ANXIOLYTIC
Includes any mention of:
- **Benzodiazepines**: Alprazolam (Xanax), diazepam (Valium), lorazepam (Ativan), clonazepam (Klonopin), temazepam (Restoril), triazolam (Halcion), chlordiazepoxide (Librium), midazolam (Versed), flunitrazepam (Rohypnol, roofies)
- **Barbiturates**: Phenobarbital, secobarbital (Seconal), pentobarbital (Nembutal), butalbital
- **Sleep medications**: Zolpidem (Ambien), zaleplon (Sonata), eszopiclone (Lunesta)
- **Other sedatives**: GHB (gamma-hydroxybutyrate, G, liquid ecstasy), carisoprodol (Soma), meprobamate
- **Behaviors**: Sedative abuse, benzodiazepine dependence, mixing benzos with other drugs, sedative withdrawal
- **Evidence**: Positive benzodiazepine test, sedative intoxication
- **Slang**: Benzos, downers, tranks, sleeping pills, nerve pills

### 7. STIMULANT
Includes any mention of:
- **Illicit stimulants**: Cocaine (powder cocaine, crack cocaine), methamphetamine (crystal meth, meth), MDMA/ecstasy/molly (when used as stimulant)
- **Prescription stimulants (when misused/abused)**: Amphetamine (Adderall, Dexedrine), methylphenidate (Ritalin, Concerta), lisdexamfetamine (Vyvanse)
- **Other stimulants**: Synthetic cathinones (bath salts, flakka), khat
- **Behaviors**: Stimulant abuse, meth addiction, cocaine use, binge use, staying up for days, tweaking
- **Evidence**: Positive cocaine test, positive methamphetamine test, positive amphetamine test (without prescription), stimulant paraphernalia (pipes, straws, mirrors)
- **Slang**: Coke, blow, crack, rock, meth, crystal, ice, crank, speed, uppers, addy, study drugs, molly, bath salts

---

## Classification Rules

For EACH substance class:

- **"Present"** = Any clear mention of substances in that class, including use, possession, positive tests, treatment for that substance, or concerns about that substance
- **"Absent"** = No mention of substances in that class

### Prescription Drug Classification

When prescription drugs are mentioned, classify them in the appropriate substance class:
- **Opioid prescriptions** (Vicodin, Percocet, OxyContin, etc.) → Opioid
- **Benzodiazepine prescriptions** (Xanax, Valium, Klonopin, etc.) → Sedative_Hypnotic_Anxiolytic
- **Stimulant prescriptions** (Adderall, Ritalin, Vyvanse, etc.) → Stimulant (only if misuse/abuse indicated)
- **Sleep medications** (Ambien, Lunesta, etc.) → Sedative_Hypnotic_Anxiolytic

### Important Notes
- A case may have MULTIPLE substance classes present - classify each independently
- If a substance is mentioned in the context of treatment/recovery, still classify as "Present"
- Tobacco/nicotine should NOT be classified under any category
- Caffeine should NOT be classified under any category
- Extract the EXACT words/phrases from the original text as evidence

---

## Output Format

Return your answer as a JSON object with classification AND evidence for each substance class:

```json
{
  "Alcohol": {"status": "Present" or "Absent", "mentions": ["exact phrase 1", "exact phrase 2"]},
  "Cannabis": {"status": "Present" or "Absent", "mentions": ["exact phrase 1"]},
  "Hallucinogen": {"status": "Present" or "Absent", "mentions": []},
  "Inhalant": {"status": "Present" or "Absent", "mentions": []},
  "Opioid": {"status": "Present" or "Absent", "mentions": []},
```

```
  "Sedative_Hypnotic_Anxiolytic": {"status": "Present" or "Absent", "mentions":
[]},
  "Stimulant": {"status": "Present" or "Absent", "mentions": []}
}
```

Rules for "mentions" field:
- If status is "Present": Include 1-5 exact quotes from the text that support the
classification
- If status is "Absent": Use empty array []
- Keep each mention brief (the specific drug term or short phrase, not entire
sentences)
- Include prescription drug names when relevant

Return ONLY the JSON object without any additional text or explanation.